\documentclass[letterpaper]{article} 
\usepackage{aaai2026}  
\usepackage{times}  
\usepackage{helvet}  
\usepackage{courier}  
\usepackage[hyphens]{url}  
\usepackage{graphicx} 
\usepackage{natbib}  
\usepackage{caption} 
\usepackage{algorithm}
\usepackage{algorithmic}
\usepackage{booktabs}
\usepackage{amsmath}
\usepackage{multirow}

\makeatletter
\protected\def\my@emoji@pic #1#2{%
  \leavevmode
  \ifvmode
    \lower\dimexpr #1\p@*1/10\relax
    \hbox{\includegraphics[height={#1\p@}]{#2}}%
  \else
    \includegraphics[height={#1\p@}]{#2}%
  \fi
}
\def\my@emoji@math #1{%
  \mathchoice
    {\my@emoji@pic\tf@size{#1}}%
    {\my@emoji@pic\tf@size{#1}}%
    {\my@emoji@pic\sf@size{#1}}%
    {\my@emoji@pic\ssf@size{#1}}%
}
\protected\def\myemoji #1{%
  \ifmmode
    \my@emoji@math{#1}%
  \else
    \my@emoji@pic\f@size{#1}%
  \fi
}
\makeatother

\newcommand{\peach}{\myemoji{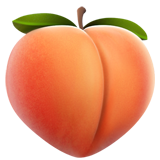}}

\newcommand{\bomb}{\myemoji{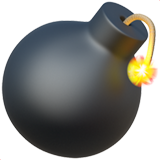}}

\newcommand{\hourglass}{\myemoji{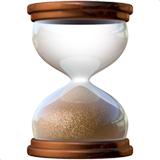}}
\newcommand{\lantern}{\myemoji{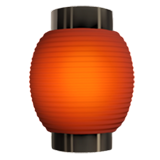}}
\newcommand{\performingarts}{\myemoji{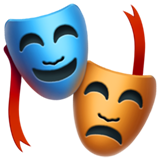}}

\usepackage{newfloat}
\usepackage{listings}
\DeclareCaptionStyle{ruled}{labelfont=normalfont,labelsep=colon,strut=off} 
\floatstyle{ruled}
\newfloat{listing}{tb}{lst}{}
\floatname{listing}{Listing}
\title{EMODIS: A Benchmark for Context-Dependent Emoji Disambiguation in Large Language Models}
\author{
    Jiacheng Huang\textsuperscript{\rm 1}\thanks{Corresponding author.},
    Ning Yu\textsuperscript{\rm 2},
    Xiaoyin Yi\textsuperscript{\rm 2}
}
\affiliations{
    \textsuperscript{\rm 1}School of Artificial Intelligence and Computer Science, Hubei Normal University, Huangshi, China\\
    \textsuperscript{\rm 2}School of Computer Science and Technology, Chongqing University of Posts and Telecommunications, Chongqing, China

    huangjc@hbnu.edu.cn,\\ \{d210201034, d200201032\}@stu.cqupt.edu.cn    
}

\begin{document}

\maketitle

\begin{abstract}
  Large language models (LLMs) are increasingly deployed in real-world communication settings, yet their ability to resolve context-dependent ambiguity remains underexplored.  
  In this work, we present EMODIS, a new benchmark for evaluating LLMs' capacity to interpret ambiguous emoji expressions under minimal but contrastive textual contexts.  
  Each instance in EMODIS comprises an ambiguous sentence containing an emoji, two distinct disambiguating contexts that lead to divergent interpretations, and a specific question that requires contextual reasoning.  
  We evaluate both open-source and API-based LLMs, and find that even the strongest models frequently fail to distinguish meanings when only subtle contextual cues are present.  
  Further analysis reveals systematic biases toward dominant interpretations and limited sensitivity to pragmatic contrast.  
  EMODIS provides a rigorous testbed for assessing contextual disambiguation, and highlights the gap in semantic reasoning between humans and LLMs.
\end{abstract}

\begin{links}
    \link{Code}{https://github.com/JiaCheng-Huang/CODIS}
\end{links}

\section{Introduction}

Large language models (LLMs)~\cite{DBLP:journals/corr/abs-2303-08774,deepseekai2025deepseekr1incentivizingreasoningcapability}, which are trained on massive corpus, have demonstrated remarkable performance across a wide range of downstream tasks, such as emotional understanding~\cite{DBLP:conf/aaai/LuCLTZ025}, content generation~\cite{sahinuc-etal-2024-systematic}, and commonsense reasoning~\cite{DBLP:conf/aaai/ZongDCLW25}.
Since LLMs continue to advance rapidly, comprehensive and rigorous evaluation of their capabilities has become increasingly essential, particularly in understanding nuanced aspects of natural language.

Due to the ambiguity of natural language, expressions with multiple meanings can be easily misunderstood by LLMs and even humans in the absence of sufficient context.
Such ambiguity has also been intensified in digital communication where emojis, despite enriching human's modes of expression, often carry multiple interpretations that heavily depend on context, increased potential for misunderstanding.
For instance, consider the sentence ``She just sent me a $\peach$ last night''.
In isolation, the emoji $\peach$ can evoke multiple interpretations, that is, it may represent a literal peach, or serve as a slang reference to buttocks or sexual innuendo.
However, when additional context is provided, the intended meaning becomes significantly clearer. 
Specifically, when the preceding conversation involves a discussion about summer fruits, the emoji is likely to be interpreted literally.
In contrast, if the context includes flirtatious or suggestive exchanges, it tends to be understood figuratively.

Although semantic disambiguation has been a widely studied topic, and several benchmarks have been proposed to evaluate LLMs in this regard, there is currently no benchmark specifically designed to assess the ability of LLMs to resolve emoji-related ambiguities in context-dependent scenarios.
Table~\ref{tab:benchmark_comparison} summarizes recent benchmarks for LLMs.
This limitation means existing benchmarks cannot assess the ability of LLMs to understand sentences with emoji in a context-dependent manner.

\begin{figure}[htbp]
  \centering
  \includegraphics[width=8.4cm]{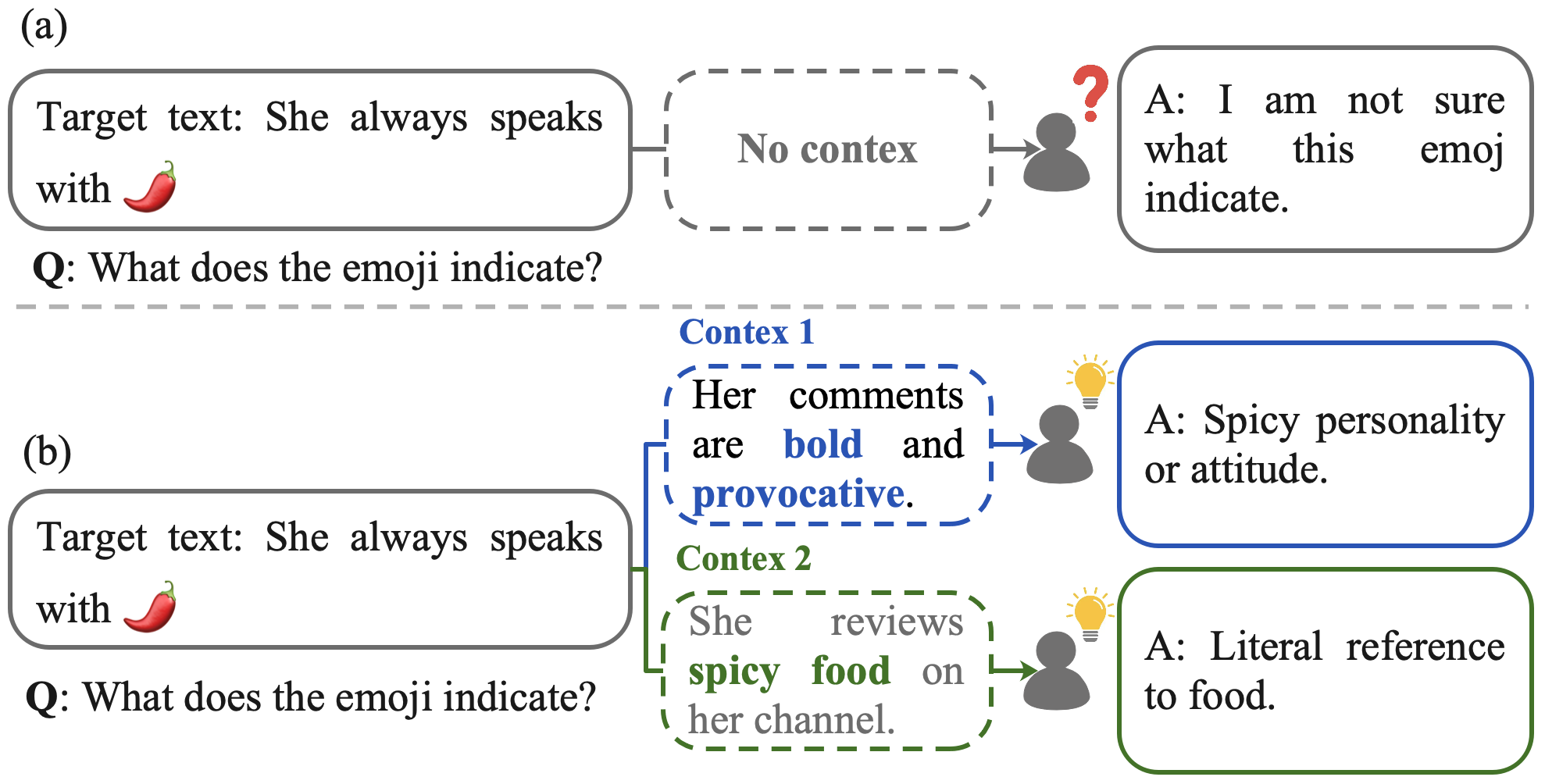}
  \caption{\label{fig:schematic}An illustration of our benchmark. Interpretation of sentence with emoji can be significantly influenced by contextual information. }
\end{figure}

\begin{table*}[htbp]
  \centering
    \begin{tabular}{lccc}
      \toprule
      \textbf{Benchmark}                                          & \textbf{Target Task}      & \textbf{Answer Format}  & \textbf{Evaluator} \\ \midrule
      DiBiMT\citep{campolungo-etal-2022-dibimt} & Word Sense Disambiguation               & Open-form text    & Metrics        \\
      GLADIS\citep{chen-etal-2023-gladis}            & Acronym Disambiguation          & Open-form text          & Metrics  \\
      ZELDA\citep{milich-akbik-2023-zelda}          & Entity Disambiguation            & Span-based linking      & Metrics  \\
      CVTE-Poly\citep{DBLP:conf/interspeech/ZhangTLWZ023} & Polyphone Disambiguation    &  Multiple-choice           & Metrics        \\
      AmbiQT\citep{bhaskar-etal-2023-benchmarking}               & Text-to-SQL Parsing     & Structured text       & GPT        \\
      CHAmbi\citep{zhang-etal-2024-chambi}        & Chinese Ambiguity Challenge          & Multiple-choice         & GPT     \\ \midrule
      \textbf{EMODIS (Ours)}               & Emoji Disambiguation & Open-form text & GPT     \\ \bottomrule
      \end{tabular}
  \caption{Comparison of proposed EMODIS with recent benchmarks addressing semantic or contextual ambiguity.}
  \label{tab:benchmark_comparison}
  \end{table*}

To address this challenge we introduce a new benchmark named EMODIS, which is designed to test the ability of LLMs in \textbf{EMO}ji \textbf{DIS}ambiguation of context-dependent sentences.
As shown in Figure~\ref{fig:schematic}, we adopt a question-answering format inspired by existing semantic disambiguation benchmarks~\cite{luo-etal-2024-codis} and our benchmark distinguishes itself from prior works in following aspects:
first, each target sentence contains an emoji whose meaning is inherently ambiguous without additional context;
second, the questions are intentionally designed to spotlight these ambiguities, requiring additional context for accurate interpretation;
third, for every sentence-question pair, EMODIS provides two subtly different contexts, each leading to a distinct interpretation of the emoji.
All sentences, contexts, questions, and answers are manually curated to ensure high quality and linguistic diversity.
Our evaluation of several widely used large language models on EMODIS reveals a significant gap between model performance and that of humans in resolving emoji ambiguity through contextual cues.   
Further analysis shows that these models often fail to leverage subtle contextual differences that shift the emoji's meaning, underscoring the necessity for improved context-aware semantic comprehension in LLMs.

Our contributions are summarized as follows:
\begin{itemize}
  \item We construct a high-quality dataset of 1000 manually curated instances. Each instance includes an ambiguous emoji-containing sentence, a disambiguation question, and two carefully designed contrastive contexts, each supporting a distinct plausible interpretation.
  \item We conduct extensive evaluations across both API-based and open-source LLMs, revealing significant gaps in context sensitivity, over-reliance on prior associations, and inconsistent performance across context types.
  \item We provide detailed analysis such as context sensitivity and interpretation bias, highlighting the core limitations of current LLMs in context-driven semantic interpretation.
  \end{itemize}

\section{Related Work}

\subsection{Context in Semantic Ambiguities}
Context plays a pivotal role in resolving semantic ambiguity, particularly when the interpretation of a sentence cannot be determined in isolation.
Traditional semantic disambiguation tasks have primarily focused on word-level distinctions, such as Word Sense Disambiguation (WSD), where the goal is to assign a correct sense label to a target word given its local context~\cite{DBLP:conf/aaai/ZhangL25a,kruk-etal-2024-silent}.
While effective in lexical disambiguation, this formulation largely neglects the complexities that arise when ambiguity exists across the entire sentence.

In real-world communication, especially in informal and digital contexts, ambiguity often extends beyond the lexical level. 
A sentence may contain figurative language, implied references, or symbols such as emojis, whose interpretation relies not only on syntactic clues but also on the broader conversational or situational context. 
For example, metaphoric or sarcastic expressions can yield drastically different meanings depending on prior discourse, speaker intention, or cultural knowledge. 
Such ambiguities cannot be resolved by analyzing isolated tokens, but instead requires models to integrate extra contextual information.

Recent studies emphasize the importance of context-aware modeling across various tasks~\cite{zhang-etal-2019-incorporating,zhao-etal-2024-large,luo-etal-2024-codis}, showing that model performance in such scenarios declines significantly when deprived of contextual information. 
However, sentence-level semantic disambiguation remains underrepresented in existing benchmarks, particularly in scenarios involving non-standard linguistic elements like emojis or symbolic expressions.

\subsection{Semantic Disambiguation Benchmarks}

Existing benchmarks for semantic disambiguation focus on various ambiguity types in natural language. 
DiBiMT~\cite{campolungo-etal-2022-dibimt} studies sense ambiguity in machine translation outputs, while GLADIS~\cite{chen-etal-2023-gladis} targets acronym disambiguation across domains. 
ZELDA~\cite{milich-akbik-2023-zelda} evaluates zero-shot entity disambiguation using minimal contextual cues. 
AmbiQT~\cite{bhaskar-etal-2023-benchmarking} focuses on query intent ambiguity in text-to-SQL tasks. 
In the Chinese context, CHAmbi~\cite{zhang-etal-2024-chambi} provides a fine-grained taxonomy of ambiguous expressions, including figurative and syntactic cases. 
Other works have explored clarification-based interaction~\cite{zhang-etal-2024-clamber} and semantic obfuscation under adversarial inputs~\cite{xiao-etal-2024-toxicloakcn}, reflecting the growing complexity of ambiguity in modern NLP scenarios.

However, current benchmarks primarily operate in textual settings and rarely consider symbolic elements like emojis, which are prevalent in digital communication and often context-dependent in meaning. 
These symbolic ambiguities are not well captured by existing resources, leaving a gap in evaluating large language models' ability to resolve emoji-related ambiguity. 
To address this, we introduce EMODIS, a benchmark that focuses on disambiguating context-dependent sentences with emoji using contextual cues, providing a new perspective on context-sensitive semantic understanding.

\section{EMODIS}
EMODIS is proposed for evaluating the capability of LLMs in resolving context-dependent sentences with emoji. 
Inspired by~\citet{luo-etal-2024-codis}, we adopt a contrastive design where each query appears in two variants, each accompanied by a distinct context leading to different interpretations. 
In this section, we present the overall task formulation, dataset construction process, evaluation protocol, and data statistics.

\subsection{Taxonomy of Context}

\begin{figure*}[ht]
  \centering
  \includegraphics[width=13.9cm]{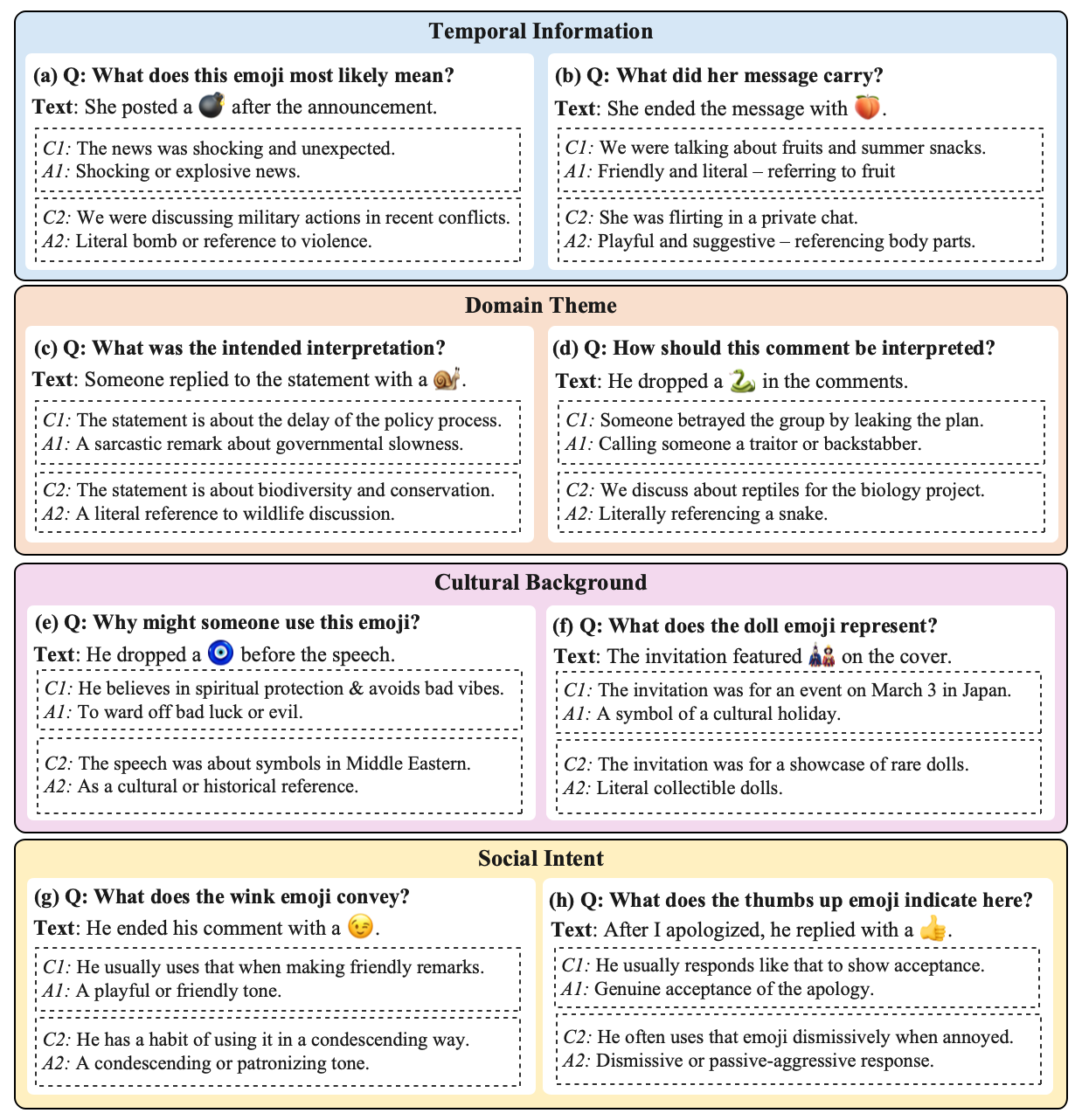}
  \caption{\label{fig:taxonomy} Taxonomy of our EMODIS benchmark. For each category, we provide two representative cases. Each case consists of a target sentence, a question, and two contrasting contexts that lead to different answers. Questions (Q), contexts (C), and answers (A) are labeled accordingly to highlight the disambiguation task. }
\end{figure*}

Given the complexity and diversity of natural language communication, cataloging all possible forms of context that contribute to disambiguating symbolic expressions is inherently challenging. In designing EMODIS as a benchmark for evaluating context-sensitive disambiguation, we focus on how the question posed to the model defines the nature of the required contextual reasoning. Rather than categorizing examples solely by the emoji involved, we categorize them by the type of interpretive challenge that the question sets up, as resolved through contrasting contexts.
Specifically, we identify four representative types of disambiguating context, each corresponding to a distinct kind of reasoning required by the question, as illustrated in Figure~\ref{fig:taxonomy}.

\paragraph{Temporal information.} Interpretation depends closely on the timing and sequence of events described by context. 
As Figure~\ref{fig:taxonomy}(a) shows, a symbol following a surprising announcement is interpreted as explosive news, whereas in a military discussion it points to a literal weapon. 
Similarly, Figure~\ref{fig:taxonomy}(b) illustrates how casual conversation about snacks leads to a literal reading, while private exchanges suggest a more intimate or playful tone.

\paragraph{Domain theme.} Recognizing the topical field implied by context is crucial for resolving ambiguity. 
In Figure~\ref{fig:taxonomy}(c), reference to policy delay prompts a sarcastic interpretation, while biodiversity context leads to a literal reading about wildlife. 
Figure~\ref{fig:taxonomy}(d) presents how domain framing distinguishes between figurative and literal meanings depending on whether the discussion involves betrayal or biology.

\paragraph{Cultural background.} Cultural knowledge provides essential clues for disambiguation, as shown in Figure~\ref{fig:taxonomy}(e), where a symbol evokes spiritual protection in one context, but points to cultural or historical reference in another. 
Figure~\ref{fig:taxonomy}(f) contrasts a holiday invitation with a showcase of collectibles, leading to different interpretations.

\paragraph{Social intent.} Disambiguation often relies on identifying tone and interpersonal attitude. 
Figure~\ref{fig:taxonomy}(g) shows that a symbol may signal playful friendliness in one context, but a patronizing tone in another. 
In Figure~\ref{fig:taxonomy}(h), the same gesture indicates either sincere acceptance or dismissive response, depending on the social intent.

This taxonomy ensures that EMODIS provides broad coverage of pragmatic reasoning demands, enabling fine-grained evaluation of large language models' ability to adapt interpretation to contextual cues posed by specific question types.

\begin{figure}[htbp]
  \centering
  \includegraphics[width=8.4cm]{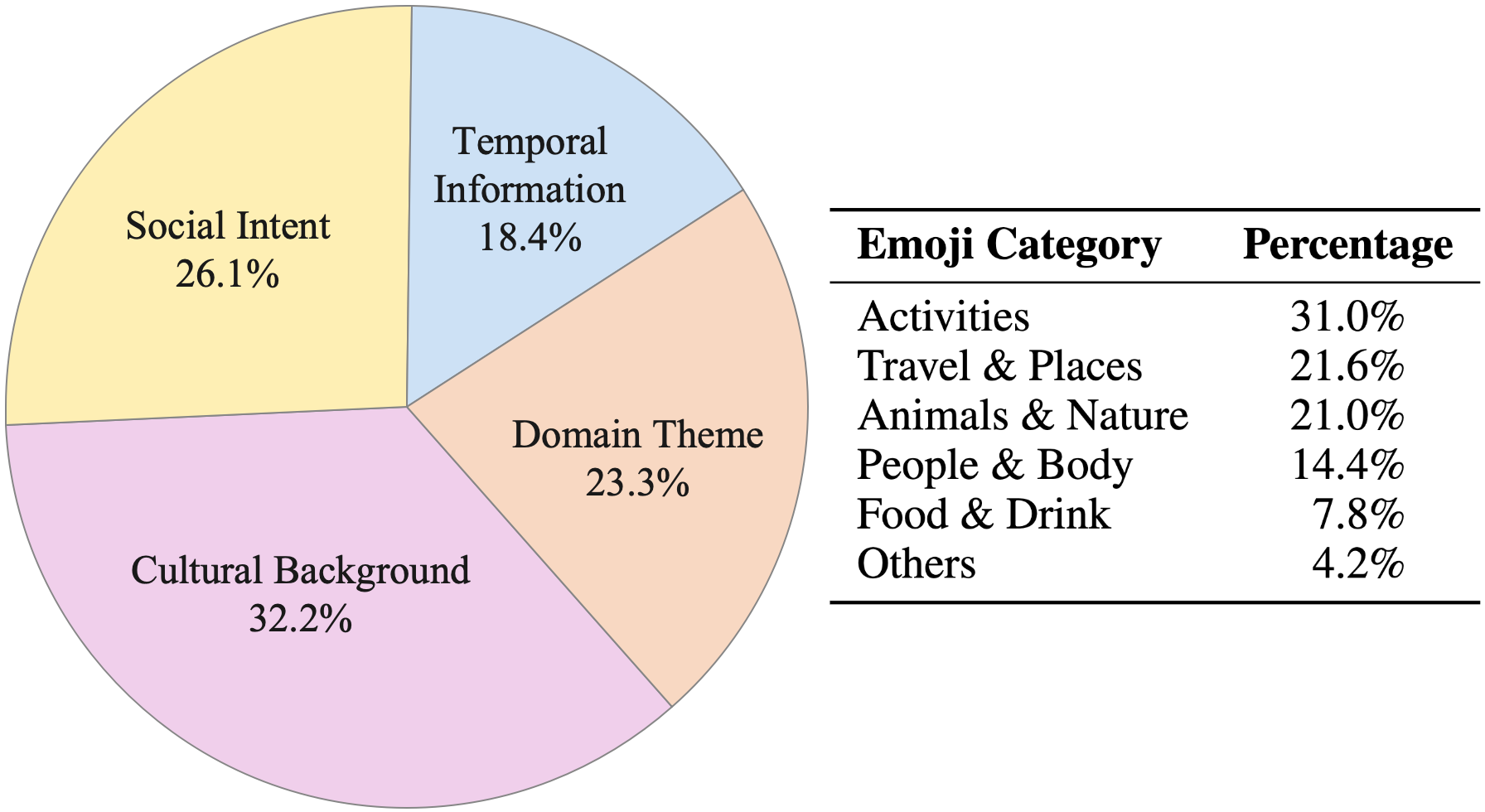}
  \caption{\label{fig:pie} Distribution of context taxonomy (left) and emoji categories (right) in our EMODIS benchmark. }
\end{figure}


\subsection{Task Formulation}
The objective of EMODIS is to test whether LLMs can distinguish between multiple interpretations of an emoji in a sentence, depending on subtle changes in context. 

To ensure that models do not guess answers without understanding the context, we structure our dataset in paired examples, denoted as $(T, C_1, Q)$ and $(T, C_2, Q)$ formally,
where $T$ refers to target text with emoji, $C_1$ and $C_2$ are two pieces of different context, and $Q$ represents question.
We input the paired examples to LLMs and obtain the outputs $O_1$ and $O_2$.
Take Figure~\ref{fig:taxonomy}(a) as an example, we provide a target text ``She posted a $\bomb$ after the announcement'' and give context that the news was shocking and unexpected in first query.
In the second query, we provide different context indicating that the discussion was about military actions in recent conflicts.  
expected outputs are that the emoji refers to shocking news or literal bomb respectively.

Besides, we apply the following constraints:
(1) Each question is neutral and under-informative by design, i.e., it cannot be answered correctly based on the sentence alone.
(2) Two contexts $C_1$ and $C_2$ are written such that they resolve the emoji's ambiguity in opposite directions.
(3) Each answer is concise and disjoint from the other.

We manually author all sentences, contexts, questions, and answers. The questions are phrased in diverse ways to avoid surface matching, and the contexts are designed to be minimal yet semantically impactful.

\subsection{Evaluation Method}

Following~\citet{fu2024mmecomprehensiveevaluationbenchmark}, we adopt pair-wise accuracy ($\mathrm{Acc}_p$) and query-wise accuracy ($\mathrm{Acc}_q$) as the evaluation metrics for our EMODIS benchmark.  
The definitions of the two evaluation metrics are as follows:
\begin{equation}
  \mathrm{Acc}_p = \frac{1}{N} \sum_{i=1}^{N} \left[ I(O_{i1},A_{i1}) \cdot I(O_{i2}, A_{i2}) \right],
\end{equation}
\begin{equation}
  \mathrm{Acc}_q = \frac{1}{2N} \sum_{i=1}^{N} \left[ I(O_{i1}, A_{i1}) + I(O_{i2}, A_{i2}) \right],
\end{equation}
where $O_{i1}$ and $O_{i2}$ are the model’s answers to the $i$-th target text under contexts $C_1$ and $C_2$, respectively, and $N$ is the number of pairs.  
$I(a,b)$ is an indicator function defined as:
\begin{equation}
I(a,b) =
\begin{cases}
1 & \text{if } a \text{ matches } b, \\
0 & \text{otherwise}.
\end{cases}
\end{equation}

\subsection{Data Collection}
In this section, we describe the process of constructing the EMODIS benchmark. Our data collection involves three stages designed to ensure both the quality and diversity of the examples while capturing realistic challenges in context-dependent disambiguation.

\paragraph{Target text authoring.}  
We begin by manually authoring target text that contain an embedded emoji whose meaning is ambiguous without additional context. 
Rather than sampling from existing corpora, we manually author target text to ensure that each emoji usage presents a genuine semantic ambiguity, where at least two distinct interpretations are plausible. 
We try to craft natural and diverse text that reflect realistic usage in digital communication across various domains, including social media, messaging, and informal writing.
To avoid introducing unintended biases, authors are instructed to refrain from including lexical hints that could resolve the emoji's meaning without the intended context.

\paragraph{Context, question and answer authoring.}  
For each target text, we manually construct two distinct contexts, $C_1$ and $C_2$, where each provides minimal but sufficient information to guide the emoji toward a different, unambiguous interpretation.
Contexts are designed to be concise, realistic, and easy to process by both human annotators and models, while avoiding artificial constructions or overly explicit cues. 
We encourage diversity in the types of context, including domain background, social intent, cultural references, and temporal or situational settings. 
Alongside the contexts, we design a disambiguation question $Q$ aimed at prompting models to focus on the semantic challenge posed by the emoji in context. 
Each question is phrased carefully to avoid redundancy with the context and to require context-sensitive reasoning, rather than simple lexical matching or frequency-based guessing.
For each pair, we provide corresponding answers $A_1$ and $A_2$. These answers are authored to be unambiguous, concise, and semantically precise, reflecting the correct interpretation of the emoji given the context. We ensure that the two answers differ meaningfully and that each aligns naturally with its associated context, so as to present a clear disambiguation signal for evaluation.

\paragraph{Verification.}  
Each data instance undergoes independent review by three human annotators. 
The data are verified with following rules: 
(1) the emoji is genuinely ambiguous in isolation; (2) the two contexts lead to distinct, clearly identifiable interpretations of the emoji; (3) the question is well-posed and cannot be answered correctly without access to the context; and (4) no unintended clues in the target text or question allow models to bypass the intended reasoning. 
Instances that do not meet these criteria are revised iteratively or discarded. 
This multi-stage process ensures that EMODIS forms a robust and reliable benchmark for context-sensitive disambiguation, with high-quality, challenging examples designed for systematic evaluation of large language models.

Ultimately, we obtained a total of 1000 disambiguation instances. 
The distribution of taxonomy types and emoji categories is illustrated in Figure~\ref{fig:pie}.

\section{Experiments}

We conduct a comprehensive evaluation to test the ability of LLMs to disambiguate emoji meanings based on context. 
Our experiments are designed to examine not only raw performance but also context sensitivity, semantic generalization, figurative reasoning, interpretive bias, and human-level comparison.

\subsection{Experiment Setup}
We perform evaluation on several popular LLMs, which are divided into two groups:
\textbf{(1) API-based models:}GPT-4~\cite{DBLP:journals/corr/abs-2303-08774}, deepseek-v3~\cite{deepseekai2025deepseekv3technicalreport}, Gemini 2.5~\cite{geminiteam2025geminifamilyhighlycapable}, and Ernie Bot 4.0;
\textbf{(2) Open-source models:}LLaMA-7B~\cite{DBLP:conf/cvpr/YeXYYHL0Z024}, Qwen-7B~\cite{bai2023qwentechnicalreport}, Vicuna-7B~\cite{vicuna}, LLaMA-13B~\cite{DBLP:conf/cvpr/LiuLLL24}, and Qwen-14B~\cite{bai2023qwentechnicalreport}.
Temperature parameter of all LLMs is set to 0.2. 

Following the evaluation setup in~\citet{fu2024mmecomprehensiveevaluationbenchmark}, we frame each data instance as a context-sensitive question-answering task. Each instance consists of a target sentence containing an ambiguous emoji, a disambiguating context, and a question that prompts the model to resolve the meaning. Models are instructed to generate a natural language answer without being shown explicit options.
We provide the detailed prompts for model inference in Appendix.

To assess whether a model's answer matches the correct interpretation, we adopt an automatic evaluation strategy using GPT-4 as a verifier. Specifically, GPT-4 is prompted to compare the model's output with the groundtruth answer and decide whether the two are semantically equivalent. If the response aligns with the correct answer, it is marked as correct; otherwise, it is marked as incorrect. This procedure is applied to both contexts in each instance to determine whether the model can adjust its answer appropriately based on contextual changes.
The prompts for GPT-4 evaluation can be found in Appendix.
  
\subsection{Overall Results}

We first report the overall performance of LLMs on our EMODIS benchmark, which evaluates their ability to resolve symbolic ambiguity in emoji-laden sentences using minimal but contrastive textual context.

As shown in Table~\ref{tab:main-results}, human annotators achieve near-ceiling performance, with $\mathrm{Acc}_p$ and $\mathrm{Acc}_q$ both above 88\%, confirming that the task is well-posed and solvable when context is properly integrated. Among all evaluated models, GPT-4 achieves the best overall performance, with a pair-wise accuracy of 58.8\% and a query-wise accuracy of 75.2\%. However, this still reflects a large gap compared to human-level understanding, indicating that even the strongest LLMs struggle with pragmatic disambiguation in symbolic expressions.
API-based models consistently outperform open-source models across all categories. 
For instance, GPT-4 reaches an overall $\mathrm{Acc}_q$ of 75.2\%, while the Qwen-14B trails behind at 58.1\%. Notably, open-source models like LLaMA-7B and Vicuna-7B perform significantly worse, with $\mathrm{Acc}_p$ below 30\%, suggesting a fundamental deficiency in context-sensitive semantic reasoning.

\begin{table*}[htbp]
  \centering
  \begin{tabular}{lcccccccccc}
    \toprule
    \multirow{2}{*}{\textbf{Model}} & \multicolumn{2}{c}{\textbf{Temporal information}} & \multicolumn{2}{c}{\textbf{Domain theme}} & \multicolumn{2}{c}{\textbf{Cultural background}} & \multicolumn{2}{c}{\textbf{Social intent}} & \multicolumn{2}{c}{\textbf{Overall}} \\
                                    & $\mathrm{Acc}_p$        & $\mathrm{Acc}_q$        & $\mathrm{Acc}_p$    & $\mathrm{Acc}_q$    & $\mathrm{Acc}_p$        & $\mathrm{Acc}_q$       & $\mathrm{Acc}_p$     & $\mathrm{Acc}_q$    & $\mathrm{Acc}_p$  & $\mathrm{Acc}_q$             \\ \midrule
    GPT-4                           & 64.5                    & 76.6                    & 58.9                & 74.2                & 59.3                    & 77.5                   & 54.2                 & 72.3                & 58.8              & 75.2               \\
    deepseek-v3                     & 36.4                    & 57.6                    &  40.3               &    60.0             &  41.6                   &     64.4               &  27.6                &  50.6               & 36.7              &    58.5           \\
    Gemini 2.5                      & 63.1                    &   76.3                  & 48.3                &    67.7             &  55.1                   &  74.1                  &     28.5             &    55.9             & 48.0              &   68.2               \\
    Ernie Bot 4.0                   & 54.2                    & 73.4                    & 54.3                &    72.5             &  56.8                   &  75.2                  &   44.7               & 66.0                & 52.5              &   71.8            \\ \midrule
    LLaMA-7B                        &   19.0                  &    43.8                 &      12.5           &      35.2           &   12.4                  &     38.7               &  11.9                & 36.4                &    13.5           &   38.2            \\
    Qwen-7B                         &   28.8                  &      53.0               &  29.6               &   55.2              &   30.4                  &     54.8               &  17.2                &    43.3             &      26.5         &      51.6         \\
    Vicuna-7B                       &  30.9                   &   53.5                  &27.4                 &  53.6               &  29.5                   &            53.2        &    22.6              &       46.7          &      27.5         &  51.7                \\
    LLaMA-13B                       & 23.9                    & 54.1                    &  21.9               &    46.8             &   22.7                  &  52.6                  &    17.2              &     46.0            &    21.3           &      49.8            \\
    Qwen-14B                        &   36.9                  &  60.8                   &  39.9               &     62.4            &  38.2                   &  59.9                  &  24.1                &     50.0            &  34.7             &   58.1               \\   \midrule
    Human                           &   88.3                  &  89.2                   &  92.7               &     94.9            &   84.9                  &   88.5                 &  89.3                &   90.4              &    88.5           &  90.6                \\ \bottomrule
    \end{tabular}
  \caption{Performances of LLMs on EMODIS benchmark. $\mathrm{Acc}_p$ and $\mathrm{Acc}_q$ are reported as percentages.}
  \label{tab:main-results}
  \end{table*}

\subsection{Context Sensitivity Analysis}
As further indicated in Table~\ref{tab:main-results}, we observe a notable gap between $\mathrm{Acc}_p$ and $\mathrm{Acc}_q$ across LLMs, with open-source models exhibiting a larger disparity than API-based ones.
Such a gap between query-wise and pair-wise accuracy highlights that many models can occasionally guess correctly but fail to switch answers when the context changes.
Therefore, to further investigate the underlying cause of this behavior, two complementary metrics namely context awareness and output variability are examined, capturing a model's ability to distinguish between different contexts and to generate context-sensitive responses.

Specifically, context awareness measures how often a model provides semantically different outputs when presented with two distinct contextual inputs for the same ambiguous sentence. 
The context awareness is computed as:

\begin{equation}
  \mathrm{Score}_{ca} = \frac{1}{n_p} \sum_{i=1}^{n_p}(O_{i1} \neq O_{i2}),
\end{equation}
where $n_p$ is the number of pairs, and $O_{i1}$ and $O_{i2}$ are the LLM's outputs for the same ambiguous text under two different contexts. 
A higher score of context awareness indicates that the model is more sensitive to context changes.
On the other hand, output variability evaluates how much the LLM's output changes when context is removed. 
It reflects the LLM's reliance on contextual information to formulate its responses. 
The output variability is defined as:
\begin{equation}
  \mathrm{Score}_{ov} =\frac{1}{n_q} \sum_{i=1}^{n_q}(O^i_{nc} \neq O^i_{c}),
\end{equation}
where $n_q$ is the number of queries for each ambiguous case, and $O^i_{c}$ and $O^i_{nc}$ are the LLM's responses with and without contextual input respectively.

\begin{table}[htbp]
  \centering
  \begin{tabular}{lcc}
    \toprule
    \multirow{2}{*}{\textbf{Model}} & \multicolumn{2}{c}{\textbf{Context sensitivity}} \\
                                    & \multicolumn{1}{l}{$\mathrm{Score}_{ca}$}    & \multicolumn{1}{l}{$\mathrm{Score}_{ov}$}   \\ \midrule
    GPT-4                           & 0.406                    & 0.481                   \\
    deepseek-v3                     & 0.245                    & 0.317                   \\
    Gemini 2.5                      & 0.375                    & 0.469                   \\
    Ernie Bot 4.0                   & 0.396                    & 0.476                   \\  \midrule
    LLaMA-7B                        & 0.254                    & 0.264                   \\
    Qwen-7B                         & 0.195                    & 0.241                   \\
    Vicuna-7B                       & 0.347                    & 0.325                   \\
    LLaMA-13B                       & 0.300                    & 0.309                   \\
    Qwen-14B                        & 0.265                    & 0.284                   \\ \midrule
    Human                           & 0.959                    & 0.491                   \\ \bottomrule
    \end{tabular}
  \caption{Performance of LLMs and humans on context sensitivity metrics. Higher values indicate better ability to distinguish and respond to contextual changes.}
  \label{tab:context-analysis}
  \end{table}

As shown in Table~\ref{tab:context-analysis}, human performance is substantially higher than all LLMs on both context awareness and output variability, confirming that current LLMs still struggle to distinguish and respond appropriately to contextual differences. 
Among the evaluated LLMs, GPT-4 and Ernie Bot 4.0 achieve the highest scores on both metrics, indicating that API-based LLMs are more sensitive to contextual variation than open-source ones. 
This is consistent with their higher $\mathrm{Acc}_p$ scores in Table~\ref{tab:main-results}, further suggesting that the ability to perceive context changes plays a critical role in solving disambiguation tasks.

\subsection{Bias in Emoji Interpretation}

To further investigate why existing LLMs underperform on EMODIS, we examine whether models tend to rely on default interpretations of emojis instead of using the provided context. 
Specifically, we analyze output bias by selecting several samples across four common emojis, i.e., $\lantern$(red lantern), $\performingarts$(performing acts), and $\hourglass$(hourglass), where both literal and figurative meanings are equally represented. 

Table~\ref{tab:bias} shows that LLMs, especially open-source ones, often prefer the more frequent, figurative meaning regardless of context. 
For instance, GPT-4 demonstrates mild bias, while LLaMA-7B exhibits strong skew toward figurative interpretations.
This supports an insight that models often default to learned priors rather than integrating contextual information.
Reducing such biases is crucial for improving robustness in context-dependent interpretation.

\begin{table}[htbp]
  \centering
  \begin{tabular}{lccc}
    \toprule
    \textbf{Model} & \textbf{\begin{tabular}[c]{@{}c@{}}Red \\ Lantern\end{tabular}} & \textbf{\begin{tabular}[c]{@{}c@{}}Performing \\ Arts\end{tabular}} & \textbf{Hourglass} \\ 
    \midrule
    Groundtruth    & 17:17 & 11:11 & 13:13 \\ 
    \midrule
    GPT-4          & 15:19 & 9:13  & 11:15 \\
    DeepSeek-v3    & 13:21 & 4:18  & 10:16 \\
    Gemini 2.5     & 14:20 & 7:15  & 9:17  \\
    Ernie Bot 4.0  & 19:15 & 1:21  & 10:16 \\
    \midrule
    LLaMA-7B       & 2:32  & 3:19  & 6:20  \\
    Qwen-7B        & 5:29  & 5:17  & 7:19  \\
    Vicuna-7B      & 8:26  & 4:18  & 5:21  \\
    LLaMA-13B      & 13:21 & 5:17  & 8:18  \\
    Qwen-14B       & 12:22 & 8:14  & 9:17  \\
    \bottomrule
  \end{tabular}
  \caption{Bias in model outputs on three ambiguous emojis. Each cell shows the count of predictions favoring the literal vs. figurative meaning.}
  \label{tab:bias}
\end{table}

\subsection{Comparison of GPT-4 and Human Evaluators}
To examine the reliability of using GPT-4 as an automatic evaluator for assessing model's answer, we compare ithe $\mathrm{Acc}_p$ and agreement between human and GPT-4 on our EMODIS benchmark.
This setup is intended to verify the reliability of GPT-4 as a proxy evaluator, without relying on full human annotation for the entire dataset.
Specifically, we define agreement as the proportion of predictions on which both human annotators and GPT-4 give the same evaluation of correctness:

\begin{equation}
  \text{Agreement} =\frac{1}{N} \sum_{i=1}^{N}(\mathrm{Eval}_\text{h}(O_i) = \mathrm{Eval}_\text{g}(O_i)),
\end{equation}
where $N$ refers to instances randomly sampled from the benchmark, and $\mathrm{Eval}_\text{h}(O_i)$ and $\mathrm{Eval}_\text{g}(O_i)$ represent the binary correctness judgments of the model output $O_i$ given by human annotators and GPT-4, respectively.

As shown in Table~\ref{tab:gpt4_human}, GPT-4 achieves over 90\% agreement with human judgments across all models, with only minor discrepancies. This demonstrates that GPT-4 is largely consistent with human evaluation in assessing context-sensitive disambiguation, and can thus serve as a reliable and scalable substitute for human annotators in EMODIS evaluation.

\begin{table}[ht]
  \centering
  \begin{tabular}{lccc}
    \toprule
    \textbf{Model} & \textbf{\begin{tabular}[c]{@{}c@{}}GPT-4\\ $\mathrm{Acc}_p$\end{tabular}} & \textbf{\begin{tabular}[c]{@{}c@{}}Human\\ $\mathrm{Acc}_p$\end{tabular}} & \textbf{Agreement(\%)} \\ \midrule
    GPT-4          & 59.6                                                                      & 60.1                                                                      & 98.5                   \\
    DeepSeek-v3    & 35.9                                                                      & 34.7                                                                      & 95.2                   \\
    Gemini 2.5     & 48.3                                                                      & 47.0                                                                      & 97.0                   \\
    Ernie Bot 4.0  & 54.2                                                                      & 52.8                                                                      & 97.8                   \\ \midrule
    LLaMA-7B       & 11.7                                                                      & 12.3                                                                      & 90.5                   \\
    Qwen-7B        & 24.4                                                                      & 23.6                                                                      & 93.2                   \\
    Vicuna-7B      & 28.2                                                                      & 27.8                                                                      & 94.0                   \\
    LLaMA-13B      & 20.6                                                                      & 21.4                                                                      & 91.7                   \\
    Qwen-14B       & 32.2                                                                      & 31.7                                                                      & 95.1                   \\ \bottomrule
  \end{tabular}
  \caption{Comparison between GPT-4 and human evaluation, including pair-wise accuracy from both evaluators and their agreement rate.}
  \label{tab:gpt4_human}
\end{table}

\section{Insights and Discussions}

Our analysis of EMODIS reveals three major limitations that large language models face when handling context-dependent emoji disambiguation. These findings not only explain the observed performance gaps but also reflect deeper issues in the models' ability to integrate and reason over context.

\paragraph{Insensitivity to Contextual Contrast.}
We observe that many models, particularly open-source ones, often produce identical or nearly identical responses when the same sentence is presented under two contrasting contexts. This suggests a phenomenon that the models are not sensitive to the subtle contrast between context variants. In such cases, the model output does not reflect the shift in meaning that a human would naturally infer. Instead of treating context as an active semantic signal that reshapes interpretation, models often regard it as peripheral. This undermines their ability to capture the kind of contrastive reasoning that EMODIS is designed to test.

\paragraph{Reliance on Prior Associations.}
Our emoji-wise bias analysis indicates that models often default to the most frequent or stereotypical interpretation of an emoji, regardless of whether the context supports it. For example, emojis such as ``peach'' or ``snake'' are consistently interpreted with their figurative, socially dominant meanings (e.g., flirtation, insult), even when the literal reading is more appropriate given the context. This reveals that model predictions are heavily influenced by associations learned during pretraining, and that context is often insufficient to override these default tendencies. This behavior shows that models do not reliably prioritize contextual cues when determining meaning, and instead fall back on prior distributions.

\paragraph{Gaps in Pragmatic Reasoning.}
While many models can answer straightforward questions correctly, they struggle with cases that involve social cues, implicit tone, or cultural inference.  Our case studies and taxonomy-based breakdowns highlight this limitation, especially for emojis whose interpretation depends on sarcasm or implicit judgment.  For such examples, model outputs often fail to align with human interpretations, even when the context clearly disambiguates the intended meaning.  This suggests that current models lack the reasoning ability necessary for a robust understanding of context-sensitive symbolic language.

\section{Conclusion}
While large language models exhibit impressive capabilities across tasks, their ability to resolve semantic ambiguity in context-sensitive settings remains limited. 
In this work, we presented EMODIS, a diagnostic benchmark for evaluating large language models on context-sensitive emoji disambiguation. By structuring each example as a sentence-question pair with two contrasting contexts, EMODIS reveals models' limitations in pragmatic inference and figurative interpretation. Experiments show that even the strongest models often fail to distinguish between literal and figurative meanings when pragmatic inference is required. We further showed that interpretive bias and context neglect are common failure modes, particularly in open models. Although the benchmark is limited in scale and modality, we hope this work provides a new lens for analyzing contextual semantics in LLMs and motivates further research into pragmatic reasoning and symbolic disambiguation.

\paragraph{Limitation.}
There are some limitations in our current work. 
First, although EMODIS provides diverse textual contexts, each is deliberately kept brief to match current LLM capabilities in leveraging contextual information.    
Future versions can incorporate more complex and layered contexts to better reflect real-world usage.    
Second, while we categorize contexts into four representative types, this taxonomy does not exhaust all disambiguation factors such as speaker identity, emotional tone, or multimodal information are not considered.    
Third, our benchmark currently relies on manual construction and human verification for high quality, which limits scalability.    
In future work, we aim to expand the dataset using semi-automatic generation and explore additional modalities and context dimensions that may affect emoji interpretation.



\bibliography{aaai2026}

\section{Appendix}
This appendix provides the full prompt templates used in our experiments for both model inference and answer verification. 
These prompts are carefully designed to control for stylistic variance, encourage concise responses, and enforce consistent evaluation criteria across different models and settings.

\subsection{Prompt for Model Inference}
\subsubsection{With Context}
To assess LLM's ability to leverage contextual information for emoji disambiguation, we present the following prompt format during inference:\\
\textit{Instruction: I'll give you a text with Emoji and some additional context, which provides information closely related to the text. Please answer my question based on the text and the context. Please answer in a single word or phrase.}\\
\textit{Context: [Context Here]}\\
\textit{Sentence: [Target Text Here]}\\
\textit{Question: [Question Here]}

\subsubsection{Without Context}
To measure model behavior in the absence of contextual cues, we use the following variant of the prompt, where the sentence and question are presented without supporting information:\\
\textit{Instruction: I will provide a sentence that contains an emoji, without any additional context. Please interpret the meaning of the emoji as it appears in the sentence. Answer with a short phrase.}\\
\textit{Sentence: [Target Text Here]}\\
\textit{Question: [Question Here]}

\subsection{Prompt for GPT-4 Evaluation}
To evaluate whether the model's answer is semantically consistent with the groundtruth, we instruct GPT-4 to serve as a reference-matching verifier. 
The evaluation prompt ensures that GPT-4 focuses on semantic alignment rather than superficial differences such as formatting or phrasing:\\
\textit{Instruction: Please evaluate the output of models based on the given question and groundtruth and tell me whether the output is right. The answer is right if it follows the question in meaning and is consistent with the groundtruth. 
If you think the answer is correct according to the groundtruth, please output “right”, otherwise output “wrong”. 
You can only print “right” or “wrong” and nothing else. 
Do not be too strict about the answer. Format different from the groundtruth and minor grammar issues are allowed.}\\
\textit{Context: [Context Here]}\\
\textit{Sentence: [Target Text Here]}\\
\textit{Question: [Question Here]}\\
\textit{Model's Answer: [Model's Output Here]}\\
\textit{Groundtruth: [Groundtruth of the Question Here]}
\end{document}